# An Automated Auto-encoder Correlation-based Health-Monitoring and Prognostic Method for Machine Bearings

**Ramin M. Hasani, Guodong Wang and Radu Grosu**
Cyber-Physical Systems Group, Vienna University of Technology, Austria
ramin.hasani, guodong.wang, radu.grosu@tuwien.ac.at

## Abstract

This paper studies an intelligent ultimate technique for health-monitoring and prognostic of common rotary machine components, particularly bearings. During a run-to-failure experiment, rich unsupervised features from vibration sensory data are extracted by a trained sparse auto-encoder. Then, the correlation of the extracted attributes of the initial samples (presumably healthy at the beginning of the test) with the succeeding samples is calculated and passed through a moving-average filter. The normalized output is named auto-encoder correlation-based (AEC) rate which stands for an informative attribute of the system depicting its health status and precisely identifying the degradation starting point. We show that AEC technique well-generalizes in several run-to-failure tests. AEC collects rich unsupervised features form the vibration data fully autonomous. We demonstrate the superiority of the AEC over many other state-of-the-art approaches for the health monitoring and prognostic of machine bearings.

## 1 Introduction

In almost all industries health management of machines is notably essential. A key subsidiary of health management is condition-based monitoring (CBM) where one prognoses abnormal status of a machine based on extracted features from a group of implemented sensors and parameters. The CBM procedure therefore includes two steps; 1) Feature extraction during a run-to-failure experiment and two 2) Data processing for predicting the degradation starting point and monitoring the defect propagation during the test.

Quite a large number of methods have been proposed for the prognostic of key machine components. In many cases, handcrafted time- and frequency-domain features are derived from the sensors mounted on the machine and are used directly or post-processed by numerous methods in order to predict the condition of the machine component [Lee *et al.*, 2014]. Recently, on the other hand, artificial intelligence (AI) solutions have been vastly utilized in fault classification and condition monitoring [Jia *et al.*, 2016; Sun *et al.*, 2016; Thirukovalluru *et al.*, 2016]. AI techniques significantly enhance the quality of the feature extraction and data processing. However, they have not yet realized a fully-automated method without the use of prior knowledge for health monitoring tests.

The main attributes of an ideal health-condition monitoring and prognostic method are described as follows: The method is able to collect unsupervised useful features from the available sensory data, regardless of its size, autonomously. Such data-driven procedure should enable the user to perform monitoring in both online and offline tests, while providing an intelligent trend on the health status of the system and precisely pointing out the degradation starting point. It is highly desirable that the method automatically combines the steps of the CBM process and be entirely human-labor independent. Moreover, the ideal approach is universal and can be feasibly applied to the prognostic of various key machine components such as bearings, gears and spindles.

In the present study, we propose a novel prognostic method for machine bearings, as a key machine component which satisfies the main characteristics of an ideal CBM method. The technique is called auto-encoder-correlation-based (AEC) prognostic algorithm. We train a sparse auto-encoder for extracting unsupervised features from collected sensory data in several test-to-failure experiments and correspondingly, compute the Pearson correlation of the extracted features of the initial samples, with the upcoming samples. The output is then passes through a moving average (MA) filter. AEC algorithm then normalizes the output of the filter and accurately illustrates the health condition of the system. We evaluate the performance of our algorithm over several run-to-failure tests of machine bearings and prove its superiority in finding the degradation starting point compared to the existing methods.

Accordingly, the paper is structured as follows. Section 2, describes the-state of-the-art methods employed for the CBM and prognostic. In Section 3, we delineate the design procedure of the AEC algorithm while recapitulating the sparse auto-encoder architecture and describing the rest of the method. In Section 4, we perform various experiments with AEC and represent our results correspondingly. Within Section 4, we qualitatively and qualitatively compare our method to the other approaches. We conclude our report in Section 5.

## 2 Related Works

Comprehensive reviews on the useful prognostic methods for industrial machines including bearings have been proposed [Jardine *et al.*, 2006; Lee *et al.*, 2014]. Traditionally, time domain statistical features such as root mean squared (RMS), Kurtosis, Spectral Kurtosis and so on, have been utilised for monitoring the status of the machine key components [Qiu *et al.*, 2006]. However such features fail to provide useful information on the status of the system in some cases and therefore are not generalizable. In this regard, selection of the proper statistical features which contain useful information, is a challenge which is addressed in many health monitoring methods [Yu, 2012a; Yu, 2012b]. Such approaches lack automation as well as generalization.

Moreover, unsupervised feature extraction techniques such as principle component analysis (PCA)-based methods [He *et al.*, 2007] followed by a post-processing stage such as hidden Markov model (HMM) for health degradation monitoring provided a reasonable prediction on the status of the system [Yu, 2012a]. Despite their attractive level of accuracy, such methods are also not automated and their generalization ability has not yet clearly shown. Moreover, combination of the frequency-domain feature extraction methods such as wavelet packet decomposition with HMM are also employed in the calculation of the remaining useful life (RUL) of the key machine components [Tobon-Mejia *et al.*, 2012]. Tobon et al. proposed a useful trend on the RUL and successfully tested it on few bearing run-to-failure experiments. The approach is however, subjected to modifications to be utilized in the other test-cases [Tobon-Mejia *et al.*, 2012].

Kalman filter (KF) is also used in condition monitoring [Reuben and Mba, 2014; Wang *et al.*, 2016]. Reuben et al. proposed a method based on switching KF which can be simultaneously employed in the diagnostic and prognostic of the data [Reuben and Mba, 2014]. Wang et al. used an enhanced KF with expectation-maximization algorithm for providing accurate estimations on the health condition of the machines [Wang *et al.*, 2016]. Such methods lack the fully automation property and are subjected to manual preprocessing.

Data-driven approaches using AI techniques, have brought great advancements to the diagnostic and health condition monitoring of key machine components. Various architectures of neural networks in particular auto-encoders, are successfully employed for precise-classification of faults into different subgroups, by using pre-processed statistical time or frequency domain features and in general features extracted using prior knowledge [Yang *et al.*, 2002; Jia *et al.*, 2016; Sun *et al.*, 2016; Thirukovalluru *et al.*, 2016]. Although proposed methods are not fully autonomous, they are less human-labor dependent. However, such methods have not been yet utilized in the prognostic and condition monitoring fully autonomous. Here, for the first time, we train a sparse auto-encoder directly over the vibration data and compute the correlation of the extracted features and therefore, provide a comprehensive prognostic and health monitoring method which its performance is superior compared to many of the remarked approaches.

## 3 Auto-encoder Correlation-based (AEC) Fault Prognostic Method

In this section we describe the mechanism of the automated AEC general fault prognostic method during the run-to-failure test of machine bearings. Figure 1 graphically illustrates the structure of the AEC. An autoencoder network is trained by using the Vibration data-samples. The auto-encoder generates rich nonlinear features from the sensory data for each sample. Afterwards, the correlation coefficient matrix of the past samples until the current sample, is computed. The correlation coefficients of the features extracted for the samples generated at the beginning of the run-to-failure process, with the other available samples is normalized and correspondingly filtered by means of a moving average (MA) filter. This provides an output rate which can predicts the status of the system at each sampling time step. In the followings, we sketch our design procedures, in details.

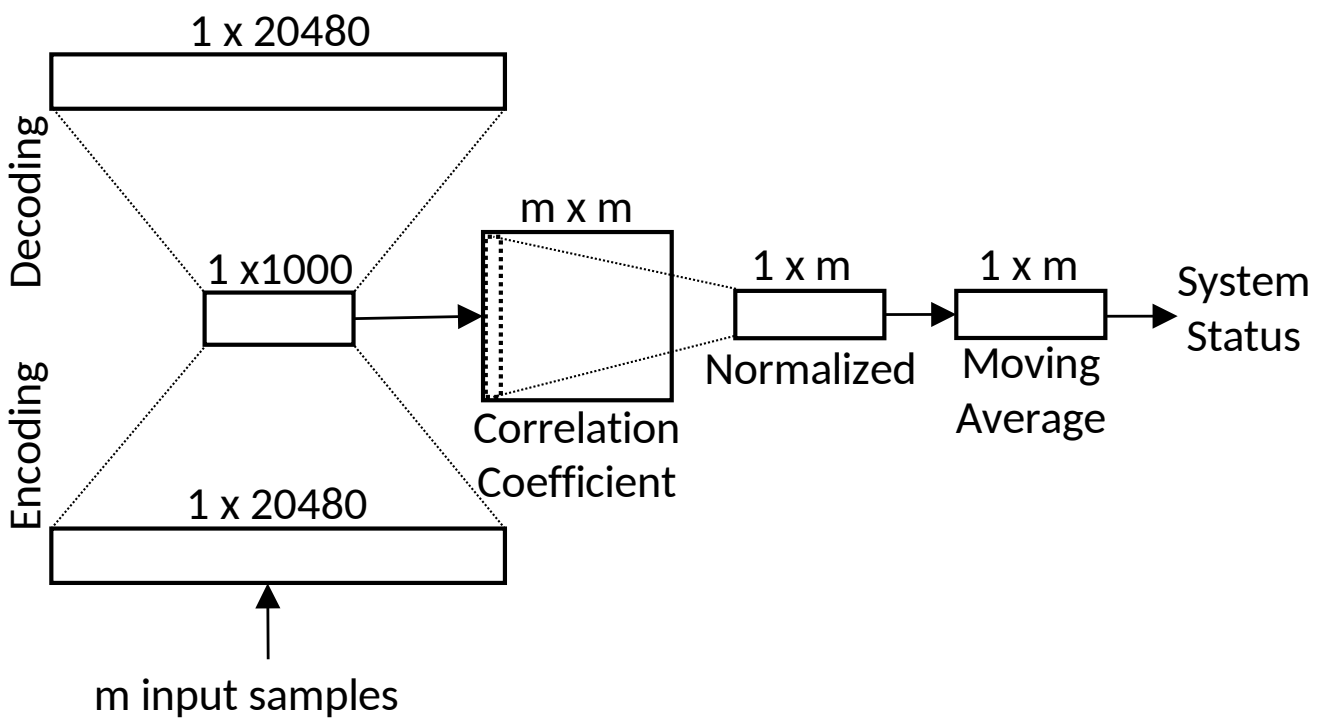

Figure 1: The architecture of the AEC fault prognostic method.

### 3.1 Sparse Auto-encoder Revisit

An auto-encoder [Boureau *et al.*, 2008], tries to learn an abstract of the identity function, to estimate the same input patterns at its output. One can place constraints on the network by limiting the size of the hidden layer and presumably discover attractive features form the data.

Let us define $x \in \mathbb{R}^{D_x}$, a $D_x$-dimension input to the auto-encoder; the encoder initially maps $x$ to a lower dimension vector $z \in \mathbb{R}^{D}$, and correspondingly generates an estimation of $x$, $\hat{x} \in \mathbb{R}^{D}$:

$$z = f(Wx + b_1); \;\; \hat{x} = f(W^T z + b_2), \qquad (1)$$

where $f$, in our design, is a saturating linear transfer function denoted in Equation 2 , $W \in \mathbb{R}^{D_x \times D}$ stands for the weight matrix, and vectors $b_1 \in \mathbb{R}^{\mathbb{D}}$, $b_2 \in \mathbb{R}^{\mathbb{D}_x}$ represent the bias values.

$$f(z) = \begin{cases} 0, & \text{if } z \leq 0 \\ z, & \text{if } 0 < z < 1 \\ 1, & \text{if } z \geq 1 \end{cases} \qquad (2)$$

We define the following cost function to be optimized subsequently:

$$C = \underbrace{\frac{1}{N}\sum_{n=1}^{N}\sum_{i=1}^{I}(x_{in} - \hat{x}_{in})^2}_{\text{mean squared error}} + \underbrace{\lambda R_{L2}}_{L_2 \text{ regularization}} + \underbrace{\sigma R_{sparse}}_{\text{sparsity regularization}}. \quad (3)$$

The first term in $C$ is the mean squared error. The second term denotes an $L_2$ regularization applied for preventing over-fitting, where $\lambda$ is the regularization coefficient. $L_2$ regularization is computed as follows:

$$R_{L2} = \frac{1}{2}\sum_{j}^{n}\sum_{i}^{k}(w_{ji})^2, \quad (4)$$

where $n$ and $k$ are the number of samples(observations) and number of variables in the training data, respectively.

The third term in the cost function determines an sparsity regularzation with an effectiveness coefficient, $\sigma$. Let the average output activation of a neuron $i$ be denoted as:

$$\hat{\rho}_i = \frac{1}{m}\sum_{j=1}^{m} f(w_i^T x_j + b_i), \quad (5)$$

and define $\rho$ as the desired output value of a neuron, One can measure their difference by using the Kullback-Leibler divergence function described below:

$$R_{sparse} = \sum_{i=1}^{D} KL(\rho \| \hat{\rho}_i) = \sum_{i=1}^{D} \rho \log(\frac{\rho}{\hat{\rho}_i}) + (1-\rho)\log(\frac{1-\rho}{1-\hat{\rho}_i}). \quad (6)$$

KL function outputs zero when $\rho$ and $\hat{\rho}_i$ are close to each other and increases when they diverge from each other. This implies sparsity in the network and it is defined in our cost function as the sparsity regularization, $R_{sparse}$.

We then train the network by applying a scaled conjugate gradient (SCG) algorithm [Møller, 1993] using MATLAB's Neural Network Toolbox.

### 3.2 Correlation Analysis, Normalization and Filtering

For each training sample, $D$ nonlinear features are constructed by the Auto-encoder. We calculate the linear dependencies of the abstract representation of each sample by utilizing the Pearson correlation coefficient (CC) which is described bellow for two samples $A$ and $B$:

$$r(A,B) = \frac{1}{D-1}\sum_{i=1}^{D}(\frac{A_i - \mu_A}{\sigma_A})(\frac{B_i - \mu_B}{\sigma_B}), \quad (7)$$

where $\mu_A$ and $\sigma_A$ are the mean and the standard deviation of $A$, respectively, and $\mu_B$ and $\sigma_B$ are the mean and standard deviation of $B$ [Fisher, 1925]. The correlation coefficient matrix for the available samples during the run-to-failure test is calculated. The first column of the CC matrix depicts the correlation of the first sample data which is recorded at the beginning of the run-to-failure process, with the other available samples. We normalize this vector between zero and one, and define it as the criteria for predicting the faulty sample and consequently determine the health-status of the system.

We finally, smoothen the shape of the output by passing it through a moving average filter. The filter is designed as follows:

$$\hat{m} = \frac{1}{w_{size}}(y(n) + y(n-1) + \cdots + y(n-(w_{size}-1))). \quad (8)$$

For the sample data $y$, the filter slides a sample-window of length $w_{size}$, over the data, and calculates the average of the covered data in each window.

## 4 Experiments with AEC

In this section we evaluate the performance of our fault prognostic method by employing it in several run-to-failure experiments on bearings. We initially introduce the dataset contents together with the objective of the tests, and illustrate the performance of the AEC method in various run-to-failure tests. We finally benchmark our results with the state-of-the-art methods in fault prognostic of the bearing machines.

### 4.1 IMS Bearing Data Set from PCoE NASA Datasets

We use the bearing dataset provided by the the Center for Intelligent Maintenance Systems (IMS), University of Cincinnati [Lee *et al.*, 2007], collected from the Prognostics Data Repository of NASA [PCoE, Accessed 2016].

The bearing test rig is shown in Figure 2. A shaft is coupled to a AC motor and is rotating at 2000 $RPM$ while a 6000 $lbs$ load is installed on it.

Four force-lubricated bearings are mounted on the shaft. Accelerometers with high sensitivity are placed for each bearing for recording the vibrations (the position of the sensors are shown in the Figure 2). Three test-to-failure experiments are performed independently. In such tests, failures usually happened at the end of the test [Qiu *et al.*, 2006].

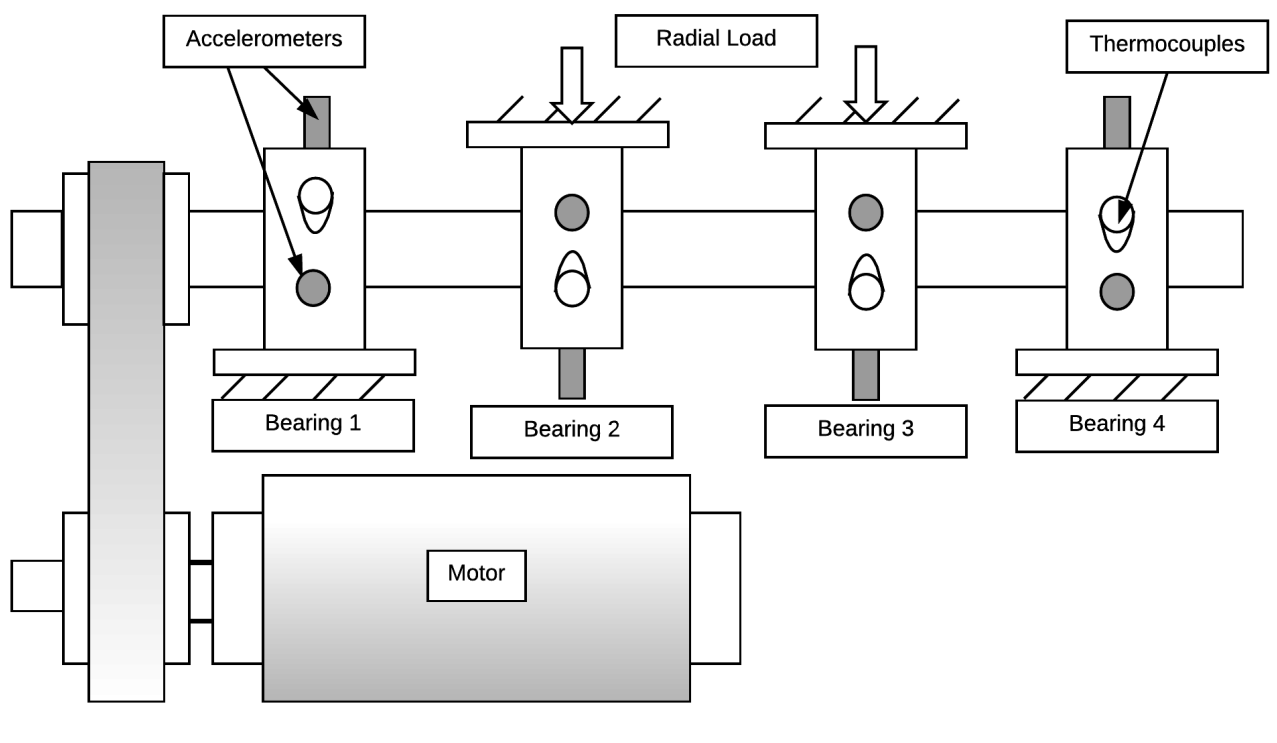

Figure 2: Bearing test implementation [Qiu *et al.*, 2006]. Four bearings are set-up on a shaft and vibration data is collected in three run-to-failure tests.

In the first experiment, two accelerometers are utilized for each bearing while in the second and third experiments one accelerometer is used. Datasets contain 1-second recordings from the accelerometers with a sampling frequency of $20KHz$, every 10 $min$, during the run-to-failure tests [Qiu *et al.*, 2006]. Table 1 represents the properties of the collected data in each experiment.

Table 1: IMS Bearing tests specification

| Experiment | # of Samples | sample size | Faulty Bearing | test-to-failure time |
|---|---|---|---|---|
| S1 | 2156 | $4 \times 20480$ | B3 and B4 | 35 days |
| S2 | 984 | $4 \times 20480$ | B1 | 8 days |
| S3 | 4448 | $4 \times 20480$ | B3 | 31 days |

For our simulations, we only train the AE network on the faulty bearings therefore we will have four different simulated experiment cases as follows: 1) Dataset 1 bearing 3 (S1B3), 2)Dataset 1 Bearing 4 (S1B4), 3) Dataset 2 bearing 1 (S2B1) and 4) Dataset 3 Bearing 3 (S3B3).

## 4.2 Results

We train the auto-encoder and consequently implement our prognostic method in MATLAB due to its compatibility with most of the industrial production systems and specially online test-to-failure environmental tools. The training process is performed on a Microsoft Azure NC-Series virtual machine powered by one NVIDIA Tesla K80 GPU. We demonstrate our method's functionality in two general frameworks where in the first one we train the auto-encoder with all the available data, for each experiment, in order to monitor the status of the system. In the second framework, we train the auto-encoder with 70% of the data and keep 30% of it for testing in order to check the prediction power of the proposed method. The training time varies between 55 min and 80 min for each case, depending on the number of input samples.

Since we only consider the vibration samples of one bearing in each simulation, the input samples' dimension, under 20 kHz sampling frequency, is a 20480-length vector. We choose 1000 hidden units for the AE which enables us to extract 1000 features from each large input vector.

We then calculate the correlation coefficient matrix of the input samples (depending on the framework in which we are working on, for the first case we feed in all the available data while for the second framework we dedicate 70% of the data for training and the network has to output a prediction on the status of the system based on the previously observed samples) and normilaze the correlation coefficient of the initial sample ( which is considered to represent the healthy status of the system) with the next samples. We then filter out the output and provide a representation of the status of the system.

### Framework1 - Health-Condition Monitoring

Figure 3 illustrates the output of the AEC for the four simulations of the three run-to-failure experiments, where we monitor the recorded data. A high AEC rate corresponds to a healthy behavior of the system while a downward trend depicts starting point of an abnormal state. High AEC indicates more correlated samples with the initial healthy status of the system. AEC clearly displays the beginning of a faulty trend together with its propagation effect. It can also demonstrate a health-characteristic portfolio for the noisy dataset, S3B3 for which many other approaches have failed to provide a health-degradation characteristics. Figure 4 graphically indicates the status of the four experiments S1B3, S1B4, and S2B1 and S3B3 in parts A, B, C and D respectively. The Color bar represents the AEC rate from 0 (blue) to 1 (red). We can distinctly observe where a significant change in the rate is occurred and therefore stop the process. We call a sample abnormal, when its AEC rate is measured 90% below the recorded sample at $t - 100$. The AEC delicately indicates the start of a faulty status and represents its gradual propagation. Table 2 summarises the detection performance of our AEC method together with some of the existing data monitoring approaches, for the IMS dataset. Detection performance is defined as the sample-time (sample number) at which the algorithm notices initiation of a degradation. Therefore, an early fault-detection determines a better performance.

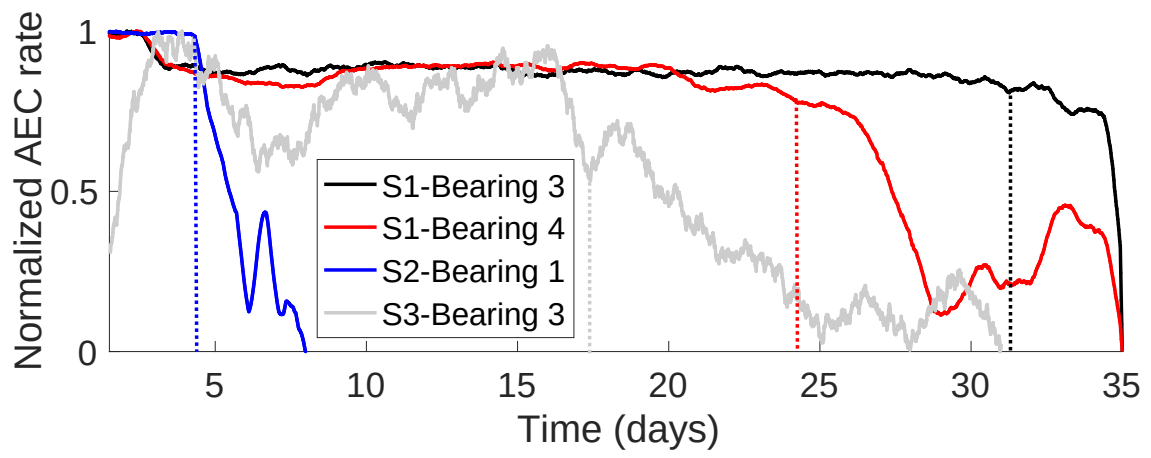

Figure 3: AEC evaluation rate for four different datasets. For every dataset, our AEC platform generates a rich trend corresponding to the status of the bearing based on the vibration sensory data. One can feasibly detect abrupt changes in the AEC rate. The vertical dot lines corresponds to the sample time after which AEC platform start decreasing dramatically. This indicates the degradation starting point.

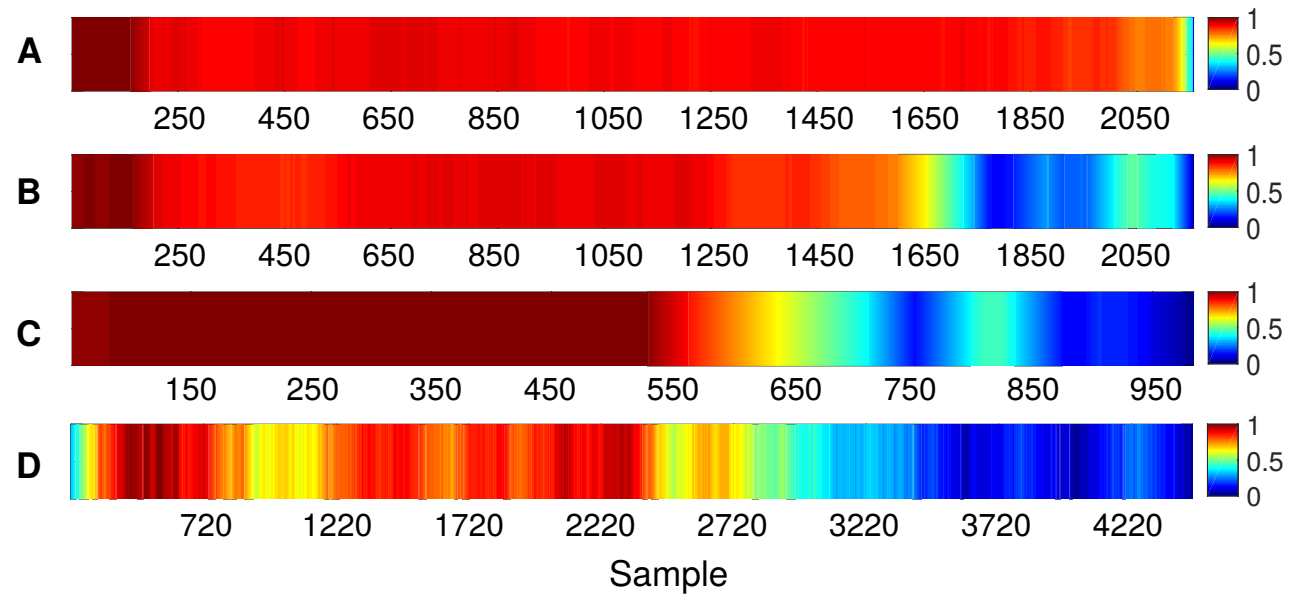

Figure 4: Visualization of the status of four different bearings in four run-to-failure experiments. The color bar represents the value of the normalized AEC rate. The higher is the rate (the more red is the color bar) the less probable it is for the system to be in a faulty condition. The horizontal axes shows the samples collected for the run-to-failure test in every experiments. A) S1B3, AEC rate recording. B) AEC for S1B4 C) AEC for S2B1. D) AEC for S3B3.

Table 2: Detection performance. HMM-DPCA: Hidden Markov model with dynamic PCA, HMM-PCA: Hidden Markov model with PCA [Yu, 2012a]. MAS-Kortusis: Moving average spectral kurtosis [Kim *et al.*, 2016]. VRCA: Variable-replacing-based contribution analysis [Yu, 2012a]. - means that the dataset has not been analyzed in the corresponding experiment

| Algorithm | S1B3 | S1B4 | S2B1 | S3B3 |
|---|---|---|---|---|
| | Degradation starting datapoint | | | |
| **AEC** | 2027 | 1641 | 547 | **2367** |
| HMM-DPCA | 2120 | 1760 | 539 | - |
| HMM-PCA | - | 1780 | 538 | - |
| RMS | 2094 | 1730 | 539 | No detection |
| MAS-Kurtosis | 1910 | 1650 | 710 | No detection |
| VRCA | - | 1727 | - | No detection |

We can observe that only AEC provides the degradation starting point for all the experiments. It is essential to mention that monitoring of the test-to-failure process of the experiment S3B3 is considerd as a hard challenge to be performed. Only few approaches can monitor its state while non provided the degradation starting point [Kim *et al.*, 2016]. AEC provides a reasonable prediction on such dataset as well.

**Framework2 - Online Prognostic**

In the second framework, as discussed, we train the auto-encoder over the first 70% of the available data in order to evaluate the prediction performance of the model in an online monitoring phase. Here, we study six cases including S1B3-sensor1, S1B3-sensor2, S1B4-sensor1, S1B4-sensor2, S2B1 and S3B3 where Figure 5A to 5F represent the predicted AEC rate in each experiment respectively. A sample is defined as the initiation of an abnormal state where its AEC rate reaches 90% of the rate of the 100 steps earlier sample. The prediction process starts from the samples collected from day 5 on, since before that the status of the system has been considered to be normal [Qiu *et al.*, 2006]. Using this definition, the degradation starting point is calculated with high level of accuracy and the propagation of fault is captured during the simulated run-to-failure test. Table 3 summarises the predicted degradation starting point together with the prediction accuracy in each experimnet. The prediction error is computed as the ratio of the deference between the predicted sample and the monitored fault starting point, to the total number of samples in each experiment.

We also compare AEC with most of the existing methods for fault monitoring and prognostic systems over several attributes. Such qualifications are determined as follows:

- Generalization: Ability of the method to provide clever results for various bearing status monitoring experiments.
- Status Monitoring: Ability of the method to provide a useful health-status trend during the test-to-failure experiment of the bearings.

Table 3: Prediction accuracy of the AEC method in different bearing run-to-failure tests

| **Experiment** | Fault starting point | Prediction Accuracy |
|---|---|---|
| S1B3-sensor1 | 2120 | 95.68% |
| S1B3-sensor2 | 2122 | 95.59% |
| S1B4-sensor1 | 1681 | 98.14% |
| S1B4-sensor2 | 1673 | 98.51% |
| S2B1 | 610 | 93.60% |
| S3B3 | 2435 | 98.47% |

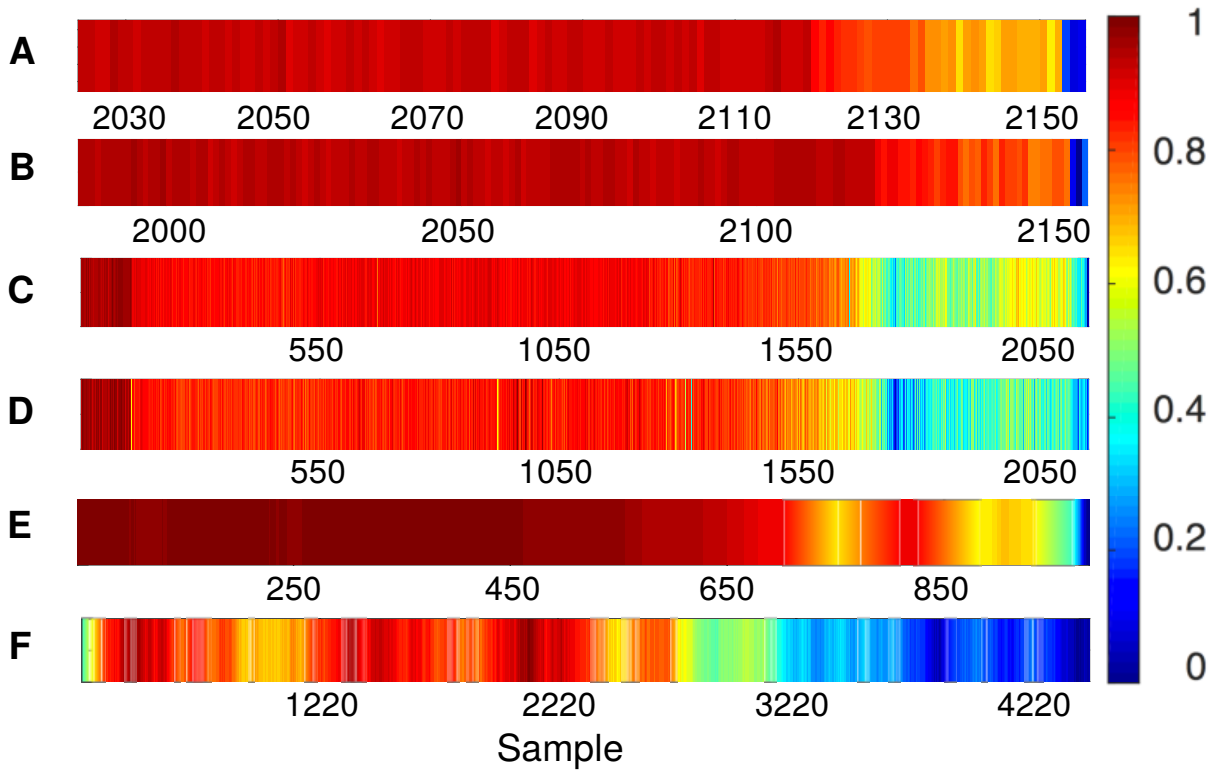

Figure 5: Online monitoring of the status of the system in different experiments: A) S1B3 first sensor B) S1B3 second sensor C) S1B4 first sensor D) S1B4 second sensor. E) S2B1 F) S3B3. The color-bar represents the AEC rate. The degradation starting point of the system in each experiment, with a high level of accuracy is predicted and AEC rate comprehensively elucidates how the fault propagates during each run-to-failure test.

- Automated: a fully autonomous fault prognostic method
- Unsupervised: Capability of the method to extract information from the sensory without any prior knowledge.
- Detection Sensitivity: Ability of the method to provide a fast-enough alert on the starting point of the degradation in the test.
- Fault-type Detection (Diagnostics): Ability of the method to detect a certain type of defect in the system and classify them into different fault classes.

Table 4 comprehensively illustrates a qualitative comparison among various fault prognostic methods utilized for bearings, based on the mentioned attributes. The assessment on the performance of each method is carefully performed based on the provided results and the detailed specifications of the methods in their corresponding report. Under such evaluations, results suggest the superiority of the AEC algorithm over existing methods where it can precisely capture the degradation initial point and provides a useful trend for the fault spread while being autonomous.

Table 4: Qualitative comparision of the performance of the existing approaches on the bearing dataset prognostic. HMM-DPCA: Hidden Markov model with dynamic PCA, HMM-PCA: Hidden Markov model with PCA [Yu, 2012a]. MAS-Kortusis: Moving average spectral kurtosis [Kim *et al.*, 2016]. VRCA: Variable-replacing-based contribution analysis [Yu, 2012a]. EET: Energy Entropy trend [Kim *et al.*, 2016]. WPSE-EMD: Wavelet packet sample Entropy [Wang *et al.*, 2011] - Empirical mode decomposition [Lei *et al.*, 2007]. Spectral-ANN: Third-order spectral + artificial neural networks [Yang *et al.*, 2002]. Fuzzy-BP: Fuzzy logic with back-propagation [Satish and Sarma, 2005]. SVM: Support Vector Machine [Yang *et al.*, 2007]. GA-SVR: Genetics algorithm-Support vector regression [Feng *et al.*, 2009]. GLR-ARMA: Generalized likelihood ratio - Autoregressive moving average [Galati *et al.*, 2008]. ++: Highly satisfies. +: Satisfies. -: The attribute is not covered -+: The attribute is fairly covered.

| **Algorithm** | Generalization | status monitoring | Automated | unsupervised | Detection sensitivity | fault-type detection |
|---|---|---|---|---|---|---|
| **AEC** | + | + | + | + | ++ | - |
| HMM-DPCA | -+ | + | - | + | + | - |
| HMM-PCA | - | + | - | + | + | - |
| RMS | - | + | - | - | -+ | - |
| Kurtosis | - | + | - | - | -+ | - |
| MAS-Kurtosis | - | + | - | - | -+ | - |
| Spectral-ANN | -+ | - | - | - | + | + |
| VRCA | -+ | + | - | - | ++ | - |
| EET | + | + | - | + | - | - |
| WPSE+EMD | - | + | - | - | + | - |
| Fuzzy-BP | -+ | - | + | - | + | + |
| SVM | - | - | - | - | + | ++ |
| GA-SVR | - | + | - | - | + | - |
| GLR-ARMA | - | + | - | - | + | -+ |

## 5 Conclusions

We introduced a new autonomous technique for fault prognostic in machine bearings. The method was based on an unsupervised feature extractions from sensors by means of an auto-encoder. Correlation analysis on the extracted features was performed and correspondingly useful trend on the status of the bearing during the test-to-failure was provided. We showed that AEC successfully monitors the status of the bearings in various experiments which confirms its generalization ability. AEC accurately predicts the degradation starting point while providing an informative trend on its propagation. Furthermore, AEC algorithm generates rich unsupervised features from the vibration data and is automated.

For the future work we intend to improve the quality of our method and apply it to the prognostic of the other key machine components such as degradation of gears, cutting tools and spindles. One can also think of a more general solution where a deep convolutional auto-encoder is designed and trained and accordingly evaluate the health-status of the system, independently, without any other follow-up steps.